\pdfoutput=1

\documentclass[journal]{IEEEtran}
\usepackage[pdftex]{graphicx}
\usepackage{float}
\usepackage{subfigure}
\usepackage{caption}
\usepackage{multirow}
\usepackage{booktabs}
\usepackage{cite}
\usepackage{amsmath,epsfig}
\usepackage{amssymb}
\usepackage{makecell}
\ifCLASSINFOpdf

\begin{document}

\title{Stereo Superpixel Segmentation Via Decoupled Dynamic Spatial-Embedding Fusion Network}

\author{
Hua Li$^*$, Junyan Liang$*$, Ruiqi Wu, Runmin Cong, Wenhui Wu, Sam Tak Wu Kwong, \textit{Fellow, IEEE}
\thanks{$^*$These authors contributed equally.
Ruiqi Wu proposed the paper idea.}
\thanks{The preliminary work of this paper was published in
ICME 2021 \cite{2021Stereo}, which is accepted as oral presentation.}
\thanks{Hua Li is with the School of Computer Science and Technology,
Hainan University,
Hainan 570228,
China (e-mail: lihua@hainanu.edu.cn).}
\thanks{Junyan Liang is with the School of Computer Science and Technology,
Hainan University,
Hainan 570228,
China (e-mail: liangjunyan@hainanu.edu.cn).}
\thanks{Ruiqi Wu is with the School of Computer Science and Artificial Intelligence,
Wuhan University of Technology,
Wuhan 430070,
China (e-mail: wuruiqi0722@gmail.com).}
\thanks{Runmin Cong is with the Institute of Information Science,
Beijing Jiaotong University,
Beijing 100044,
China,
and also with the Beijing Key Laboratory of Advanced Information Science and Network Technology,
Beijing Jiaotong University,
Beijing 100044,
China (e-mail: rmcong@bjtu.edu.cn).}
\thanks{Wenhui Wu is with the College of Electronics and Information
Engineering,
Shenzhen University,
Shenzhen 518060,
China (e-mail: wuwenhui@szu.edu.cn).}
\thanks{Sam Tak Wu Kwong is with the Department of Computer Science,
City University of Hong Kong,
Kowloon Hong Kong 999077,
China,
and also with the City University of Hong Kong Shenzhen Research Institute,
Shenzhen 518057,
China (e-mail: cssamk@cityu.edu.hk).}
}

\markboth{IEEE TRANSACTIONS ON MULTIMEDIA}%
{Shell \MakeLowercase{\textit{et al.}}: Bare Demo of IEEEtran.cls for IEEE Journals}
\maketitle

\begin{abstract}
Stereo superpixel segmentation aims at grouping the discretizing pixels into perceptual regions through left and right views more collaboratively and efficiently.
% Existing stereo superpixel segmentation algorithms  single view as input, while neglecting the correspondence between both views.
Existing superpixel segmentation algorithms mostly utilize color and spatial features as input, which may impose strong constraints on spatial information while utilizing the disparity information in terms of stereo image pairs.
To alleviate this issue, we propose a stereo superpixel segmentation method with a decoupling mechanism of spatial information in this work.
To decouple stereo disparity information and spatial information, the spatial information is temporarily removed before fusing the features of stereo image pairs, and a decoupled stereo fusion module (DSFM) is proposed to handle the stereo features alignment as well as occlusion problems.
Moreover, since the spatial information is vital to superpixel segmentation,
% 加半句为什么要在这里加回来 (Finish)
we further design a dynamic spatiality embedding module (DSEM) to re-add spatial information, and the weights of spatial information will be adaptively adjusted through the dynamic fusion (DF) mechanism in DSEM for achieving a finer segmentation. %加半句这么做的好处或者动机是什么，目前在method之前还没有具体强调这个模块的作用是什么，contribution那里就说了能提高性能，还需要具体讲下为什么能提高性能，后面对应的地方加一下。 (Finish)
Comprehensive experimental results demonstrate that our method can achieve the state-of-the-art performance on the KITTI2015 and Cityscapes datasets, and also verify the efficiency when applied in salient object detection on NJU2K dataset.
The source code will be available publicly after paper is accepted.
\end{abstract}

\begin{IEEEkeywords}
Stereo image, superpixel segmentation, stereo corresponding capturing, spatiality embedding.
\end{IEEEkeywords}
\IEEEpeerreviewmaketitle

\section{Introduction}
%%%%%%%%%%%%%%%%%%%%%%%%%%% what is superpixel %%%%%%%%%%%%%%%%%%%%%%%%%%%
\IEEEPARstart{S}{uperpixel} segmentation aims at grouping the discretizing pixels into some high-level correlative units as input primitives in a variety of subsequent computer vision tasks,
\textit{e.g.}, salient object detection \cite{2020Going,Video,cong2017co,xu2019video}, image dehazing \cite{yang2018superpixel}, image classification \cite{shi2019multiscale}, object recognition \cite{wang2014superpixel}, adversarial attack \cite{dong2020robust}.
Nowadays, dual cameras have been widely used in extensive industrial applications, such as assistant driving and mobile phones.
Compared with single images, stereo image pairs can obtain complementary information from the second viewpoint, which is beneficial to scene representation and object modeling \cite{li2021stereo}.
However, how to effectively utilize complementary and correspondence information to generate superpixels for stereo images is still a challenging task.

\begin{figure}[t!]
    \centering
    \includegraphics[width=\linewidth]{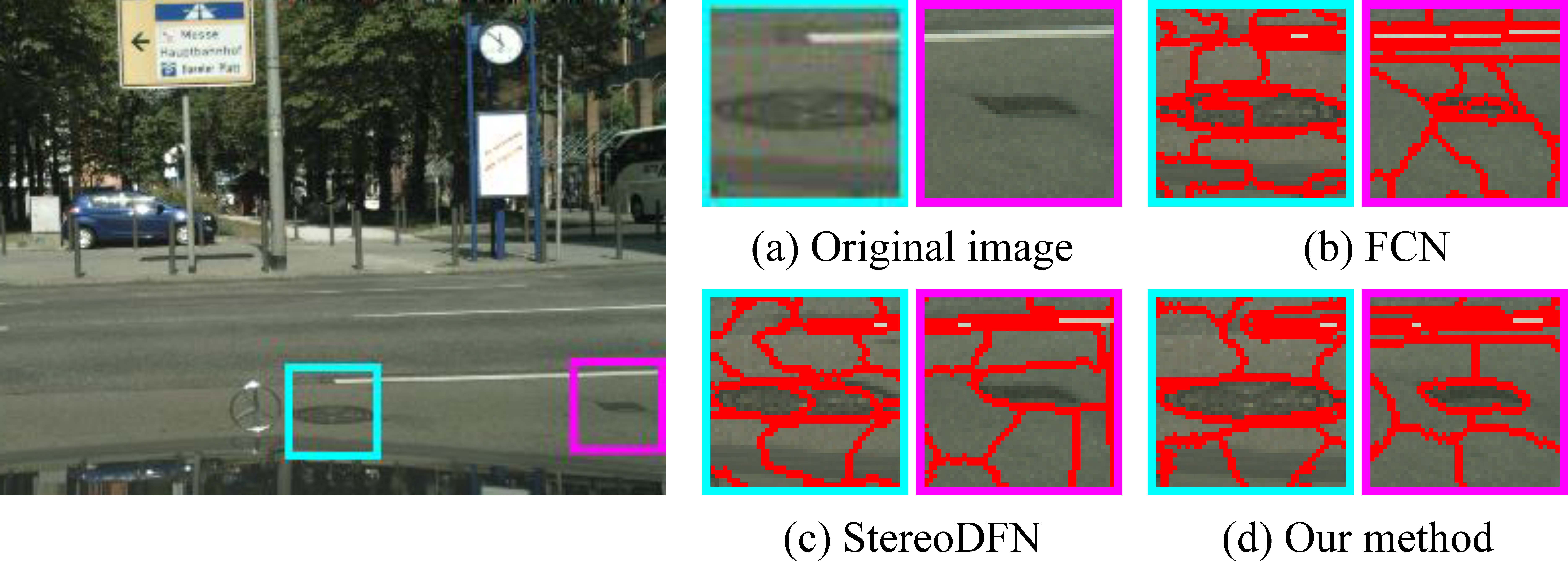}
    \caption{A simple illustration of the comparison between the state-of-the-art superpixel segmentation methods and our method.}
    \label{fig:Simple-compare}
\end{figure}
%%%%%%%%%%%%%%%%%%%%%%%%%%%%%%%%%%%%%%%%%%%%%%%%%%%%%%%%%%%%%%%%%%%%%%%%%%%
%%%%%%%%%%%%%%%%%%%%%%%%%%%%% Figure: method %%%%%%%%%%%%%%%%%%%%%%%%%%%%%%
\begin{figure*}[t]
    \centering
    \includegraphics[width=.95\textwidth]{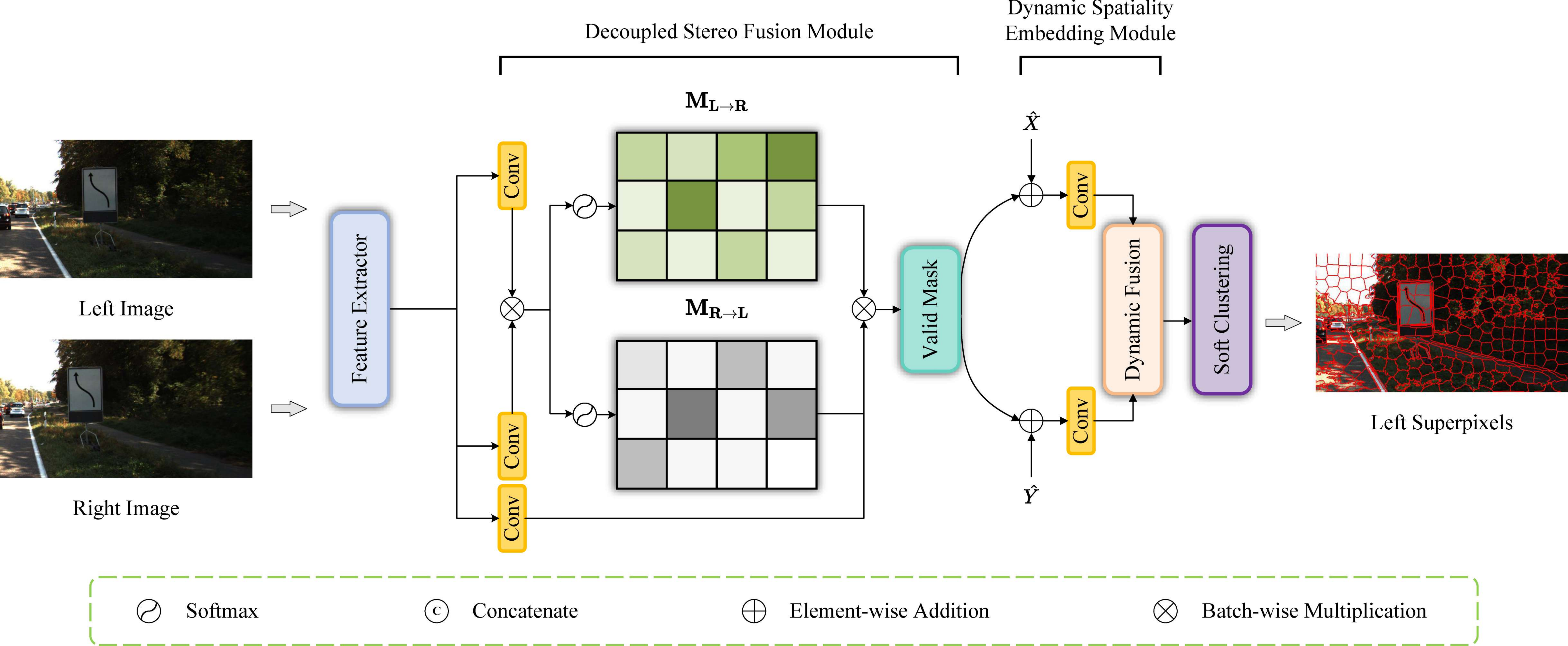}
    \caption{Overall framework of the proposed method.}
    \label{fig:method}
\end{figure*}
%%%%%%%%%%%%%%%%%%%%%%%%%%%%%%%%%%%%%%%%%%%%%%%%%%%%%%%%%%%%%%%%%%%%%%%%%%%

%%%%%%%%%%%%%%%%%%%%%%%%%%% stereo superpixel methods%%%%%%%%%%%%%%%%%%%%%%%%%%%
For stereo superpixel segmentation, stereo image pairs are generally segmented separately as single views by traditional methods. By this way, the complementarity and correlation between the left and right views of stereo image pairs are ignored and cannot be explored sufficiently \cite{cheng2015cross,liu2016complementary}. Therefore, these methods cannot be regarded as an real implementation of stereo superpixel segmentation, since the intrinsic characteristics of stereo images are neglected.
To take the collaborative relationship between left and right views into consideration,
Li \textit{et al.} \cite{li2021stereo} propose a collaborative optimization scheme to generate stereo superpixels with the parallax consistency, which is the first attempt to devise a specific superpixel segmentation method for stereo image pairs.
The method first match the corresponding regions between the left and right view of a stereo image pair. Superpixels are initialized and matched in the corresponding regions. Then, the superpixels in the left and right views are refined simultaneously via a collaborative optimization strategy.
Experimental results demonstrate it outperforms the methods that segment stereo image pairs separately.
Nevertheless, this method extracts handcrafted feature instead of deep feature by an unsupervised way, which leads to the limitation of the performance.

%%%%%%%%%%%%%%%%%%%%%%%%%%% our improvements %%%%%%%%%%%%%%%%%%%%%%%%%%%
Most recently, Wu \textit{et al.} \cite{2021Stereo} propose an end-to-end dual attention fusion network (StereoDFN) for stereo superpixel segmentation, which extracts the deep features of stereo image pairs by convolutional neural networks instead of handcrafted features. Then it models the correspondence between the left and right views via the parallax attention module to integrate the complementary information of stereo image pairs and has achieved impressive performance.
However, the existing superpixel segmentation methods like StereoDFN and superpixel sampling network (SSN) \cite{SSN} utilize the five-dimensional features (including two-dimensional spatial features XY, and three-dimensional color features Lab) as input,
which aims to extract high-dimensional deep features. For stereo superpixel segmentation task, the disparity information is often adopted to emphasize the correspondence between left and right views.
If the spatial information is input before the feature fusion, 
this may result in strong constraints on spatial information,
since the stereo features are coupled with spatial features,
which will lead to degradation of image boundaries.

The presented work significantly extends StereoDFN with a decoupling mechanism of spatial information to only use three-dimensional color features (Lab) instead of five-dimensional features (XYLab) in modeling the correspondence between both views,
thereby decoupling the stereo features (color features) and spatial features (XY) to relax the constrain of spatial information.
Considering the importance of spatial information in superpixel segmentation,
we further design a Dynamic Spatiality Embedding Module (DSEM) to re-add spatial information, the weighting of spatial information will be adaptively adjusted through the Dynamic Fusion (DF) mechanism in DSEM to fit images of different sizes, thereby obtaining a more accurate representation of spatial information and achieving a better performance.
%我们将输入的XY信息放于双目module即SFM的后面再输入，这样即能松弛在双目module对空间信息的约束，又能在超像素分割的步骤之前输入空间信息以得到更精确的分割。 (Finish)
As the simple comparison shown in Fig.~\ref{fig:Simple-compare}, we can see that our proposed method adhere to the object boundaries better than FCN \cite{FCN} and StereoDFN.

In this paper, we improve upon our previous work in ICME \cite{2021Stereo}.
The main contributions of the proposed work can be summarized as follows:
\begin{enumerate}
    \item We propose a stereo superpixel segmentation method with a decoupling mechanism of spatial information to generate superpixels for stereo image pairs, which can integrate the correspondence from both left and right viewpoints to take the stereo features alignment and occlusion problems into account.
    \item
    Since the coupling of stereo features and spatial features may impose strong constraints on spatial information while modeling the correspondence between stereo image pairs,
    we develop a spatial decoupling mechanism to model the correspondence with relaxed spatial constraint by decoupling the stereo features and spatial features, and postponing the embedding of spatial information after stereo features have been fused.
    \item We design a Dynamic Spatiality Embedding Module (DSEM) to re-add spatial information for achieving a finer segmentation. The weighting of spatial information can be adaptively adjusted via the Dynamic Fusion (DF) mechanism in DSEM to fit images of different sizes, thereby achieving a better performance.
    \item Our method achieves the state-of-the-art performance 
    compared with previous works both quantitatively and qualitatively.
    Extensive ablation studies validate the effectiveness of the proposed strategy.
    With application in salient object detection, we also demonstrate that our method can achieve superior performance in downstream task.
\end{enumerate}

The article is organized as follows. We briefly introduce the related work about existing superpixel segmentation algorithms in Section II.
Then, we propose our model and detail each key component in Section III.
The qualitative and quantitative experimental results and analyses are presented in Section IV. 
In Section V, we present the application of the proposed method in salient object detection.
Finally, Section VII concludes this article.

\section{Related Work}
The concept of “Superpixel” is first introduced in \cite{ren2003learning}, which is an over-segmentation of images and generated by grouping pixels similar in low-level properties.
Existing superpixel segmentation algorithms can be simply divided into two categories: unsupervised superpixel segmentation methods and supervised superpixel segmentation methods.
\subsection{Unsupervised Superpixel Segmentation Methods}
%unsupervised 不用讲这么多，不用再细分类了，SLIC，LSC，ASS讲下就行了，我的双目那篇也可以放进来 (Finish)
% Conventional superpixel segmentation methods can be roughly classified into two categories, including graph-based methods and clustering-based methods.
%
% Shi \textit{et al.} \cite{shi2000normalized} propose a global criterion named normalized cut (Ncut), which is one of the most representative graph-based methods. Ncut regard image segmentation as a graph partitioning problem, pixels of the images will be grouped into multiple large regions and these regions will be divided into many small units through K-means algorithm. However, Ncut is not widely used since its high computational complexity cost and inefficiencies.
%
% Topology preserved regular superpixel (TPS) \cite{tang2012topology} algorithm consider the topology and regularities of the images, which has the ability to preserve the topology of the images and generate regular superpixels.

Simple linear iterative clustering (SLIC) \cite{SLIC} is one of the most widely used unsupervised methods, which employs k-means clustering approach to generate superpixels efficiently by grouping nearby pixels based on five-dimensional color and position features of the images.
Due to SLIC has fast runtime and impressive performance, many superpixel-based applications commonly use SLIC for superpixel segmentation.
Linear spectral clustering (LSC) \cite{chen2017linear} generates superpixels based on kernel function instead of using the traditional eigen, which is not only able to produce compact superpixel, but also with low computational costs.
Considering the irregular structure of superpixels, Li \textit{et al.} \cite{li2019superpixel} propose approximately structural superpixels (ASS), they regard superpixel segmentation as a square-wise asymmetric partition problem and generate ASS by an asymmetrically square-wise superpixel segmentation way, which can preserve semantics better and largely reduces data amount.
\subsection{Supervised Superpixel Segmentation Methods}
%这个部分可以多写点，包括武睿祺的那篇也写进来 (Finish)
In recent years, inspired by the success of deep learning techniques in a wide variety of computer vision tasks, some works try to use deep learning techniques for superpixel segmentation. Jampani \textit{et al.} \cite{SSN} propose the first deep learning-based end-to-end trainable superpixel segmentation network (SSN), which is enlightened by the SLIC method. To simplify the generation of superpixels, Yang \textit{et al.} \cite{FCN} propose a lightweight fully convolutional networks (FCN) that based on encoder-decoder structure, which generates superpixels efficiently by predicting the probability map between pixels and superpixels.
More recently,
Wu \textit{et al.} propose an dual attention fusion network (StereoDFN),
they attempt to take the collaborative relationship between stereo image pairs into consideration by modeling the correspondence between them,
which is based on parallax attention mechanism.
\section{Proposed Method}
In this work, we propose a stereo superpixel segmentation method with a decoupling mechanism of spatial information,
the framework is illustrated in Fig.~\ref{fig:method}.
In general, the proposed method can be divided into the following steps:
% Feature Extractor
First, stereo image pairs with Lab color space are input
into fully convolutional network to extract the deep features.
% SFM
Then, the deep features of left and right views are fed to the Decoupled Stereo Fusion Module (DSFM),
which integrates the features from both views. 
% DSEM
Moreover, Dynamic Spatiality Embedding Module (DSEM) is proposed
to adaptively combine the spatial information with deep features.
% result
Finally, a soft clustering algorithm \cite{SSN} is adopted to generate the superpixels.

In what follows, we detail the main components of the proposed method,
which are feature extractor, Decoupled Stereo Fusion Module (DSFM),
Dynamic Spatiality Embedding Module (DSEM) and loss functions respectively.

\subsection{Feature Extractor}
\begin{figure}[t!]
    \centering
    \includegraphics[width=\linewidth]{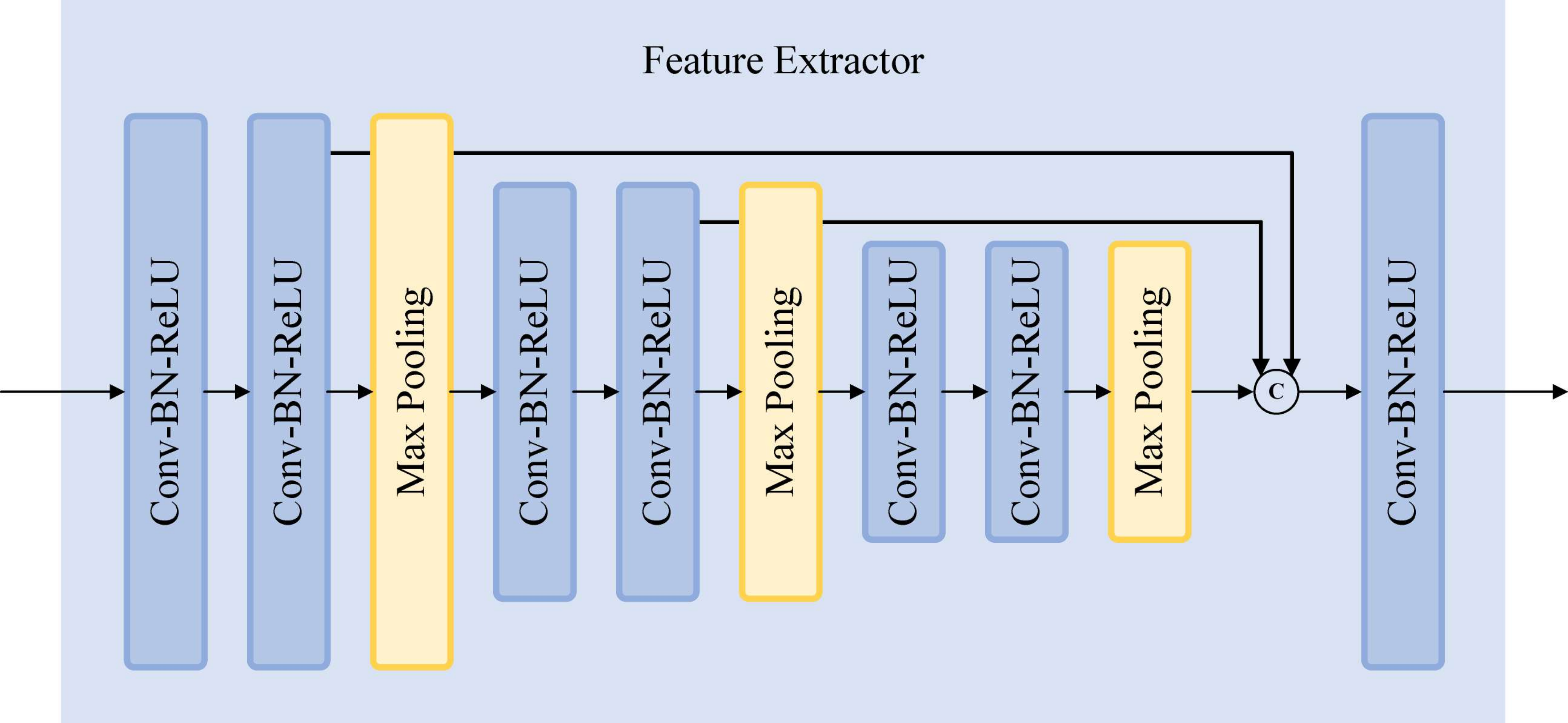}
    \caption{The schematic illustration of Feature Extractor.}
    \label{fig:feature_extractor}
\end{figure}

A pair of weight-shared Convolutional Neural Networks (CNNs) is adopted to extract the deep feature of stereo image.
The basic block is a `Conv-BN-ReLU' block, which is composed of a convolution layer with $3 \times 3$ kernel size and $64$ output channels,
a batch-normalization layer and a ReLU activation function.
Each of the two modules will be followed by a max-pooling layer for downsampling.
For features captured from Block2, Block4 and Block6,
we upsample them into the same resolution as the input image and concatenate them together.
Block7 will fuse them and generate the final output.
Through this way, the networks can effectively learn more multi-level and multi-scale features, which is benefit for both superpixel segmentation and capturing the correspondence of stereo image pairs.
The schematic illustration of feature extractor has been shown in Fig.~\ref{fig:feature_extractor}.

\subsection{Decoupled Stereo Fusion Module}
Decoupled Stereo Fusion Module (DSFM) is the key component for fusing the stereo features.
Considering the most significant problems in stereo features fusion,
such as features alignment and occlusion problem,
the proposed DSFM try to solve them via parallax attention mechanism and valid mask.

\begin{figure}[t]
	\centering
	\subfigure[Previous method.]{
        \includegraphics[width=\linewidth]{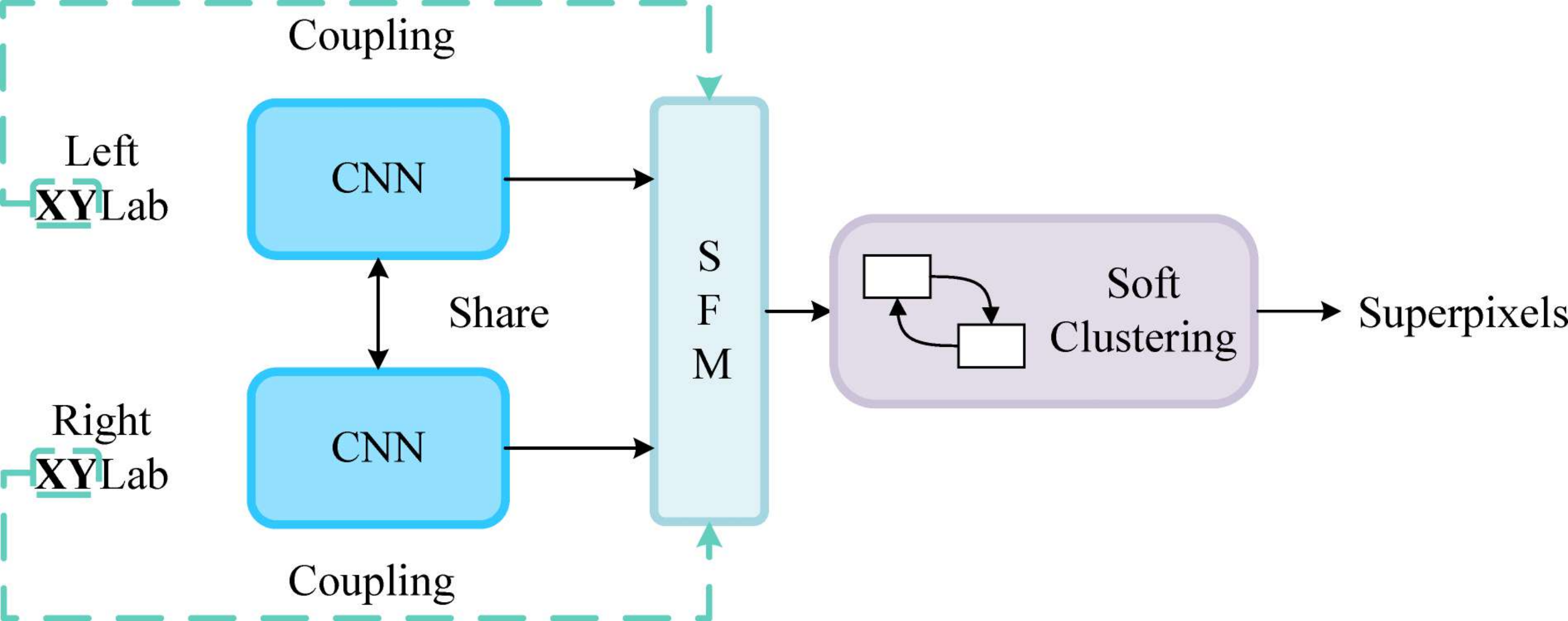}
	}
	\subfigure[Proposed method.]{
		\includegraphics[width=\linewidth]{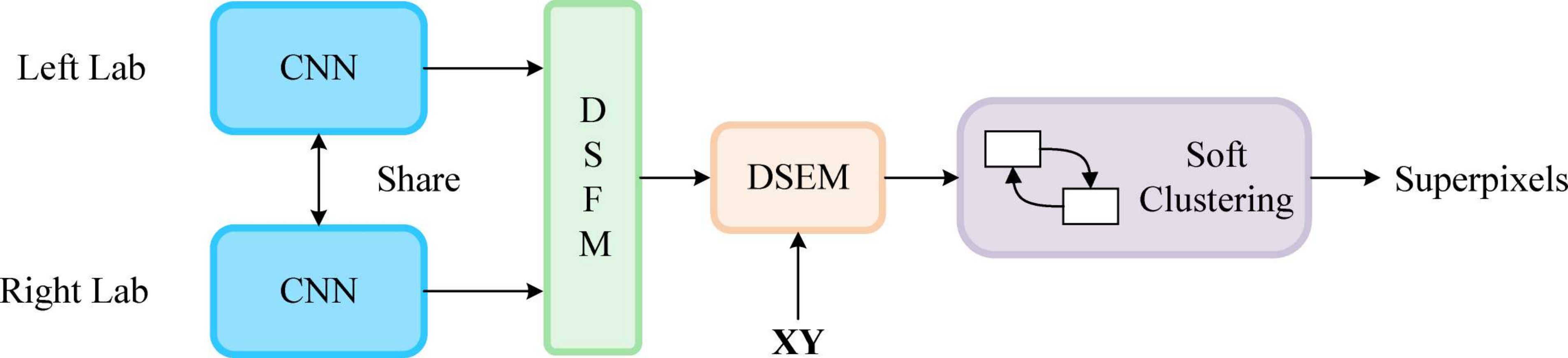}
	}
	\caption{Differences between our method and StereoDFN. 
	We only use Lab color information as input to decouple the stereo features and spatial features.
	After \textbf{D}ecoupled \textbf{S}tereo \textbf{F}usion \textbf{M}odule (DSFM),
	we design a \textbf{D}ynamic \textbf{S}patiality \textbf{E}mbedding \textbf{M}odule (DSEM) 
	to re-add spatial information.} 
	
	\label{fig:compare_previous}
\end{figure}

\begin{figure}[t] 
    \begin{minipage}[b]{.3\linewidth}
      \centering
      \centerline{\includegraphics[width=2.85cm]{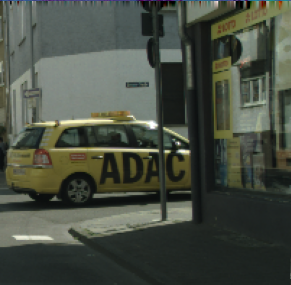}}
      \centerline{(a) Original image}
    \end{minipage}
    \hfill
    \begin{minipage}[b]{.3\linewidth}
      \centering
      \centerline{\includegraphics[width=2.85cm]{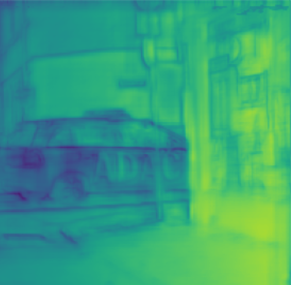}}
      \centerline{(b) Coupled}
    \end{minipage}
    \hfill
    \begin{minipage}[b]{.3\linewidth}
      \centering
      \centerline{\includegraphics[width=2.85cm]{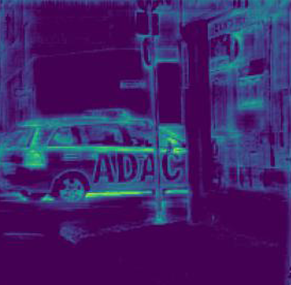}}
      \centerline{(c) Decoupled}
    \end{minipage}
    \hfill
    \caption{The visualization of deep features after modeling the correspondence between stereo image pairs.
    Note that (b) is generated by StereoDFN, while (c) is generated by our method.}
    \label{fig:decouple_visualized}
\end{figure}

\textbf{Spatial Decoupling Mechanism.}
Considering the coupling of stereo features and spatial features may impose strong constraints on spatial information while modeling the correspondence between stereo image pairs, thereby interfering with superpixels to adhere to the object boundaries.
The spatial decoupling mechanism is proposed to model the correspondence between stereo image pairs with relaxed spatial constraint.%图3和图4阐述的顺序调换一下，现讲decouple的原理，即图4，再说一个simple case show一下这样做的好处，注意这里阐述不能用实验结果的语气来讲，这里不是实验部分，只能说我们show了一个simple case直观地显示decouple的好处，然后再解释一下。(Finish)
More specifically, we remove the spatial information of input items for relaxing the constrain of spatial information.
The input of StereoDFN is a five-dimensional features (XYLab),
while the input of our proposed method is a three-dimensional features (Lab),
this is the essential difference between our proposed method and StereoDFN,
we also present the schematic illustration of the difference in Fig.~\ref{fig:compare_previous}.
Furthermore, the effectiveness of our spatial decoupling mechanism has been shown in Fig.~\ref{fig:decouple_visualized}, benefiting from decoupling stereo features and spatial features, our method eliminates the interference of spatial information on modeling, and the boundary information is much more clearly.

\textbf{Stereo Features Alignment.}
Since the corresponding pixels in stereo image pairs are located at different positions,
it is extremely difficult to fuse the stereo features directly.
Therefore, aligning the stereo features is necessary before fusion.

\begin{figure}[t]
    \centering
    \includegraphics[width=\linewidth]{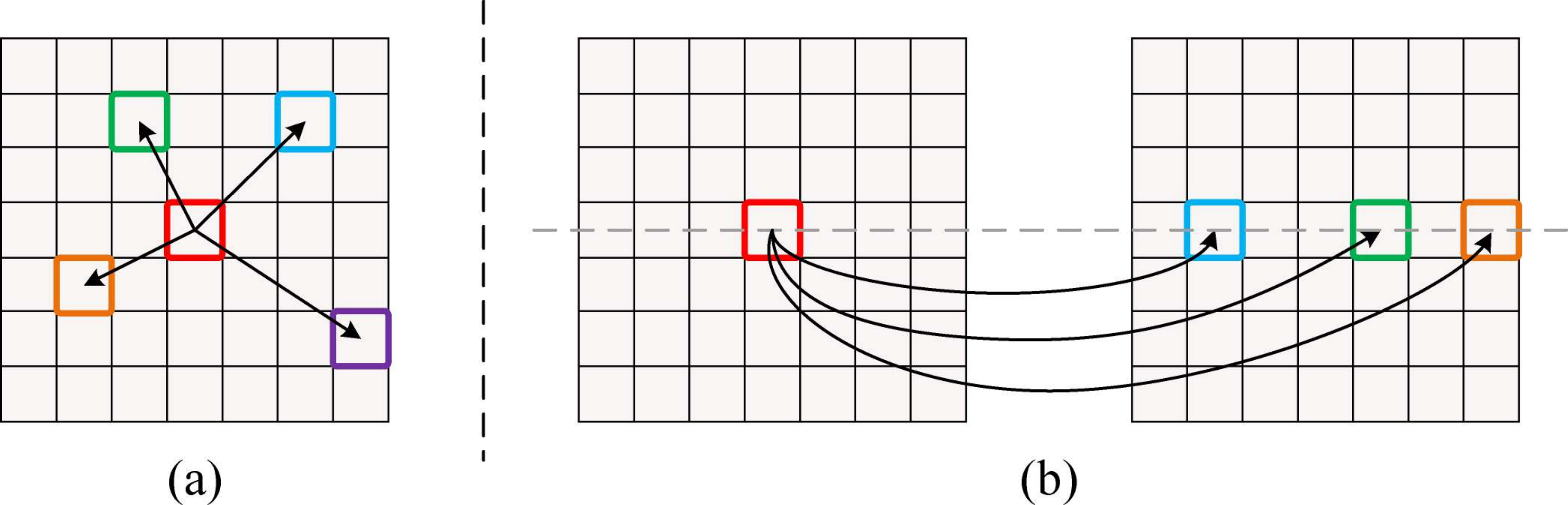}
    \caption{The schematic illustration of the difference between self attention and parallax attention mechanism.
    (a) and (b) are self attention mechanism and our parallax attention mechanism respectively.
    We can find the computational complexity of parallax-attention mechanism is much smaller
    by comparing (a) and (b).}
    \label{fig:attention}
\end{figure}
Inspired by~\cite{PAM}, the parallax attention mechanism is utilized
to model the correspondence between stereo image pairs.
% Why only consider the pixels on the same horizontal line.
Since the left and right views of the images are only translated horizontally, while aligning in the vertical direction,
the matching pixels in two views must be on the same horizontal line.
% improvement for pam
For this reason, as the schematic illustration shown in Fig.~\ref{fig:attention}(a) and Fig.~\ref{fig:attention}(b), the parallax attention mechanism is only consider the correlation of pixels on the same horizontal line, instead of all the pixels in the image like traditional self attention mechanism.
In this way, the computation complexity of attention can be largely reduced.

As the schematic illustration of aligning the features from the right view with the left view features shown in Fig.~\ref{fig:method},
for a pair of deep features $\mathcal{F}_{L}$ and $\mathcal{F}_{R}\in\mathcal{R}^{H \times W \times C}$,
we can get $A$ and $B$ from a convolution layer with $1 \times 1$ kernel size. 
Then, the parallax attention map $\mathcal{M} \in \mathcal{R}^{H \times W \times W}$ will be generated by:
\begin{equation}
    \mathcal{M}_{R \to L} = softmax(A \otimes B^T),
\end{equation}
\begin{equation}
    \mathcal{M}_{L \to R} = softmax(B \otimes A^T),
\end{equation}
where $\otimes$ denotes the batch-wise multiplication, and $T$ denotes the batch-wise transposition.
$\mathcal{M}_{L \to R}(i, j, k)$ and $\mathcal{M}_{R \to L}(i, j, k)$ represent the contribution of the position $(i, j)$ in one view to position $(i, j)$ in another view.
In this way, the supplementary information of one view can be obtained through another view, which can be formulated as follows:

\begin{equation}
    \hat{\mathcal{F}}_{L} = \mathcal{M}_{R \to L} \otimes \mathcal{F}_{right},
\end{equation}
\begin{equation}
    \hat{\mathcal{F}}_{R} = \mathcal{M}_{L \to R} \otimes \mathcal{F}_{left},
\end{equation}
where $\hat{\mathcal{F}}_{L}$ and $\hat{\mathcal{F}}_{R}$ denote the aligned features.

\textbf{Occlusion Problem.}
Occlusion always exists in stereo images due to violent disparity variation,
which will lead to an inaccurate stereo features fusion.
Therefore, we further add the occlusion handling part in DSFM to take occlusion problem into consideration.

Taking the example of handling the occlusion in the right image.
Assuming that a pixel located at $(i, j)$ is in the occlusion region,
for any $k \in [1,W]$, $\mathcal{M}_{L \to R}(i,j,k)$ is an extremely low value
since any pixel on the same horizontal line in the left view is not relevant to it.
Thus, the valid mask $O_{L \to R}$ can be generated from parallax attention map $\mathcal{M}_{L \to R}$,
which can be formulated as Eq.~(\ref{valid-mask}):
\begin{equation}
	\label{valid-mask}
	O_{L \to R}(i,j) =
	\begin{cases}
	1, &{\sum_{k \in [1, W]} \mathcal{M}_{L \to R}(i, k, j) \textgreater \tau} \\
	0, &{\sum_{k \in [1, W]} \mathcal{M}_{L \to R}(i, k, j) \leq \tau},
	\end{cases}
\end{equation}
where $\tau$ is a threshold set to $0.1$. 
% \begin{figure}[h]
%         \begin{minipage}[b]{.3\linewidth}
%           \centering
%           \centerline{\epsfig{figure=figures/left-mask.eps,width=2.8cm}}
%           \centerline{(a) Left Image}
%         \end{minipage}
%         \hfill
%         \begin{minipage}[b]{.3\linewidth}
%           \centering
%           \centerline{\epsfig{figure=figures/right-mask.eps,width=2.8cm}}
%           \centerline{(b) Right Image}
%         \end{minipage}
%         \hfill
%         \begin{minipage}[b]{.3\linewidth}
%           \centering
%           \centerline{\epsfig{figure=figures/mask.eps,width=2.8cm}}
%           \centerline{(c) Occlusion Mask}
%         \end{minipage}
%         \hfill
%         %
%         \caption{Visualization of occlusion mask. The yellow part reprensents the occlusion part in right viewpoint.}
%         \label{visual mask}
%     \end{figure}
Then, the occlusion handling part will fuse the stereo features.
The left fused features $\Tilde{\mathcal{F}}_{L}$ can be obtained as Eq.~(\ref{eq:fuse}):
\begin{equation}
    \label{eq:fuse}
    \Tilde{\mathcal{F}}_{L} = Concat(\hat{\mathcal{F}}_{L} \circ O_{L \to R} + \mathcal{F}_{L} \circ (1 - O_{L \to R}), \mathcal{F}_{L}),
\end{equation}
where $Concat(,)$ represents the concatenate operation on channel dimension,
while $\circ$ represents the Hadamard product.
Finally, $\Tilde{\mathcal{F}}_{L}$ will be fed to a `Conv-BN-ReLU' block
for reducing the channel size to the size of origin features.

\subsection{Dynamic Spatiality Embedding Module}
To prevent the spatial information influence the stereo correspondence modeling,
the spatial information has been removed in the input items.
However, spatial information is indispensable for superpixel segmentation method to adhere to object boundaries more accurately, which is vital to achieve a better performance.
Therefore, spatial information is re-added via the Dynamic Spatiality Embedding Module (DSEM) to take both conditions into consideration simultaneously, so that we can not only eliminate the disadvantage of spatial information in modeling the correspondence, but also utilize the advantage of spatial information.
DSEM consists of two parts, which are Spatiality Embedding (SE) and Dynamic Fusion (DF).
The architecture of DSEM can be seen in Fig.~\ref{fig:method}.
\begin{figure}[t]
    \centering
    \includegraphics[width=\linewidth]{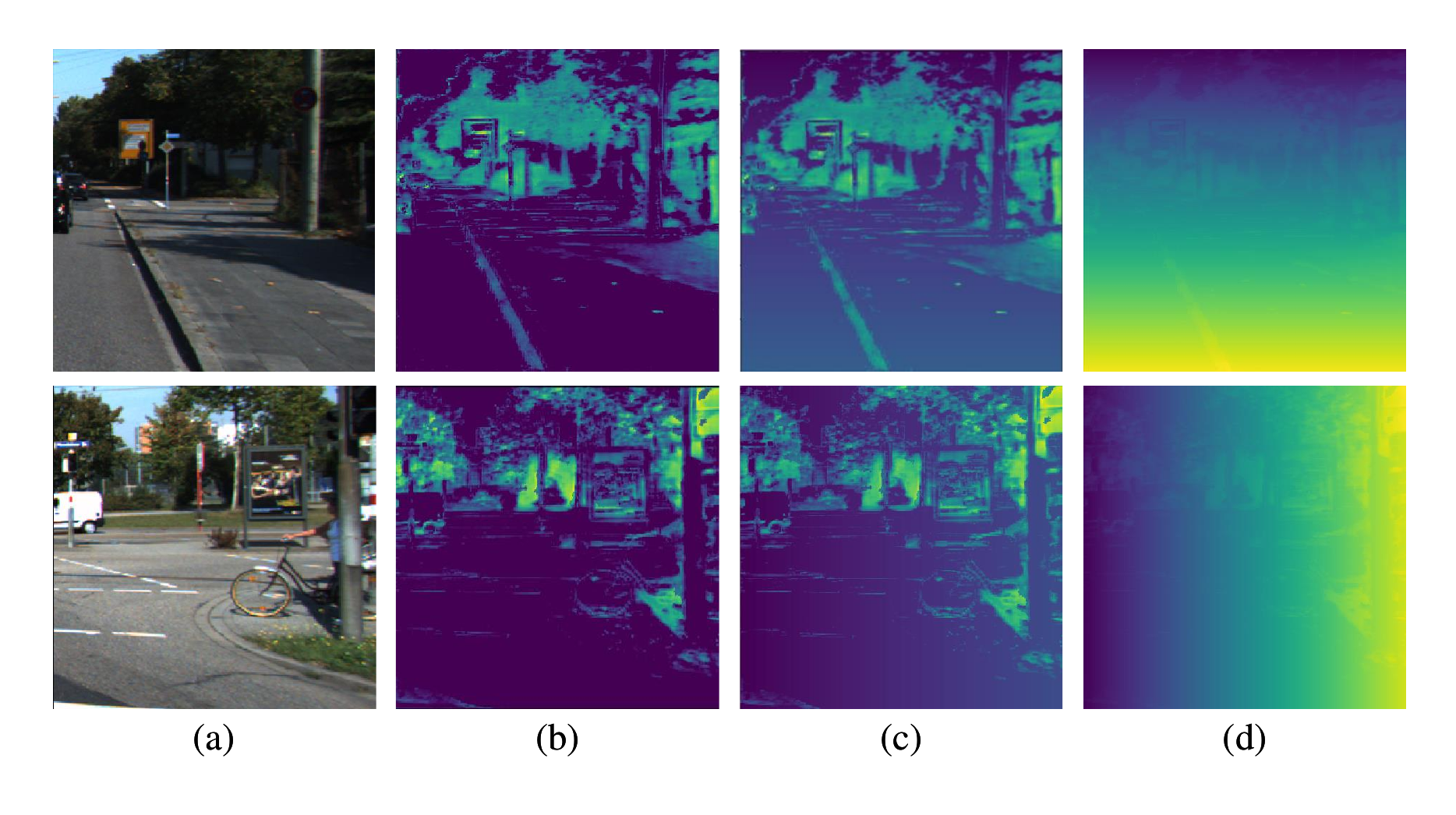}
    \caption{(a) denotes the input images with $300 \times 300$ resolution,
    (b) the deep features, 
    (c) the features after adding the spatial information with a value domain of $(0,1)$, 
    and (d) the features after adding the spatial information zoomed in the manner of~\cite{SLIC, SSN}.
    We can see our embedding strategy preserves more details than (d).}
    \label{fig:coordinate}
\end{figure}

\textbf{Spatiality Embedding.}
%We use the element-wise addition to add spatial information in our module.
A reliable superpixel segmentation algorithms require the ability to handle images with different resolutions.
However, the value of spatial information can be extremely large for a high-resolution image,
which will pollute the image feature representation if the spatial information is embedded directly.
In order to avoid such a disadvantage,
we normalize the spatial information $X$ and $Y$ as Eq.~(\ref{eq:range}):
\begin{equation}
    \label{eq:range}
    \hat{X} = \frac{X}{max(X)},~\hat{Y} = \frac{Y}{max(Y)},
\end{equation}
where $X$ and $Y$ are spatial information on horizontal and vertical direction, respectively.

After normalizing, $\hat{X}$ and $\hat{Y}$ is added to fused features and get $\Tilde{\mathcal{F}_{X}}, \Tilde{\mathcal{F}_{Y}}$.
Then, a convolution layer with $1 \times 1$ kernel size is followed to embed the spatial information.
% The output channel number of it is set to $16$, which is smaller than the input channel number.
In this way, we can prevent an over-consideration of spatial information.
Finally, $\Tilde{\mathcal{F}_{X}},~\Tilde{\mathcal{F}_{Y}}$ is concatenated with input features and send them to dynamic fusion part.
Fig.~\ref{fig:coordinate} shows the effectiveness of our embedding strategy.

\begin{figure}[t!]
    \centering
    \includegraphics[width=\linewidth]{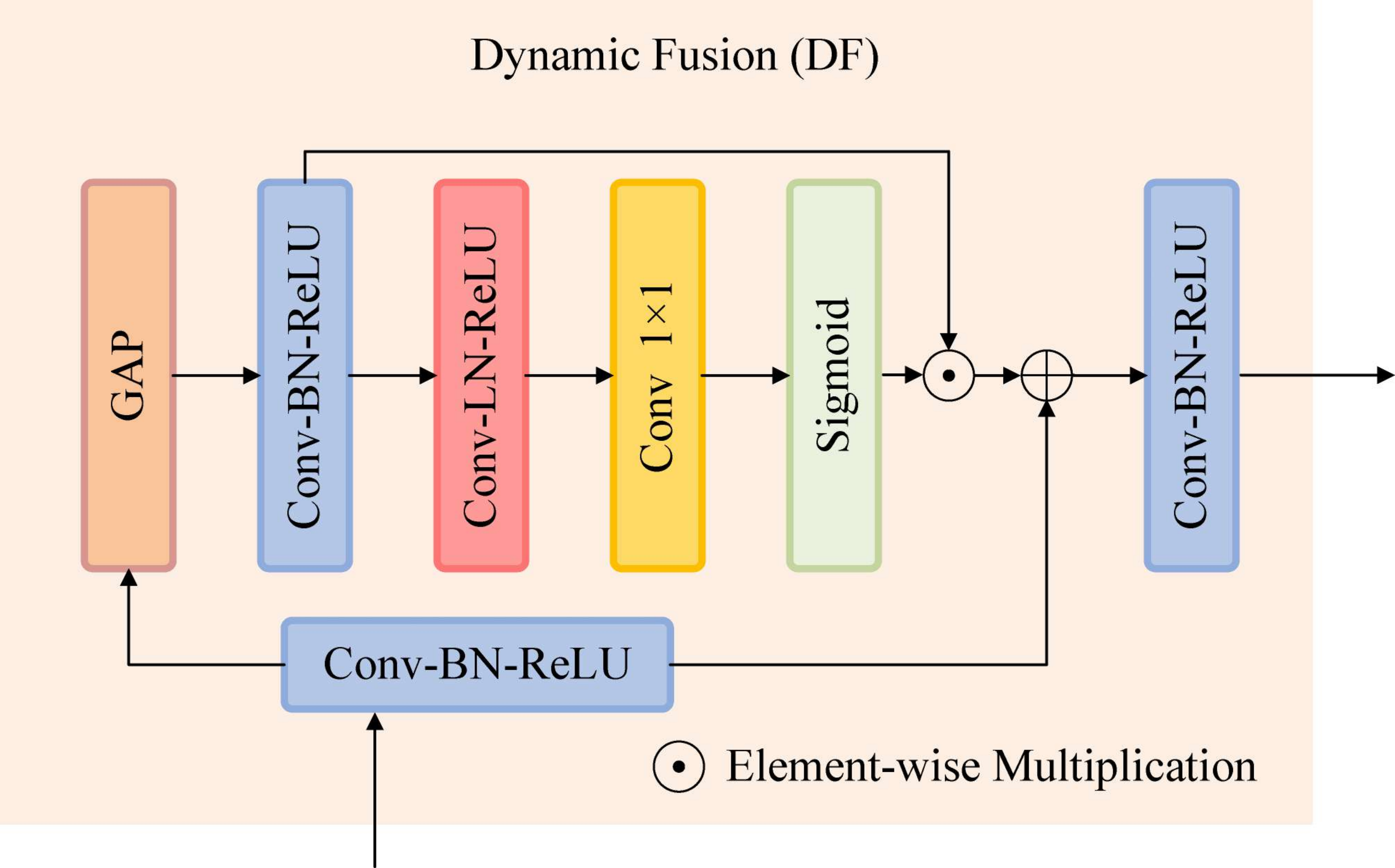}
    \caption{The schematic illustration of Dynamic Fusion mechanism.}
    \label{fig:DF}
\end{figure}
%%%%%%%%%%%%%%%%%%%%%%%%%% Figure: ASA/UE/BR %%%%%%%%%%%%%%%%%%%%%%%%%%
\begin{figure*}[!t]
    \centering
    \subfigure[Performance on the KITTI2015 dataset.]{
        \includegraphics[width=.32\textwidth]{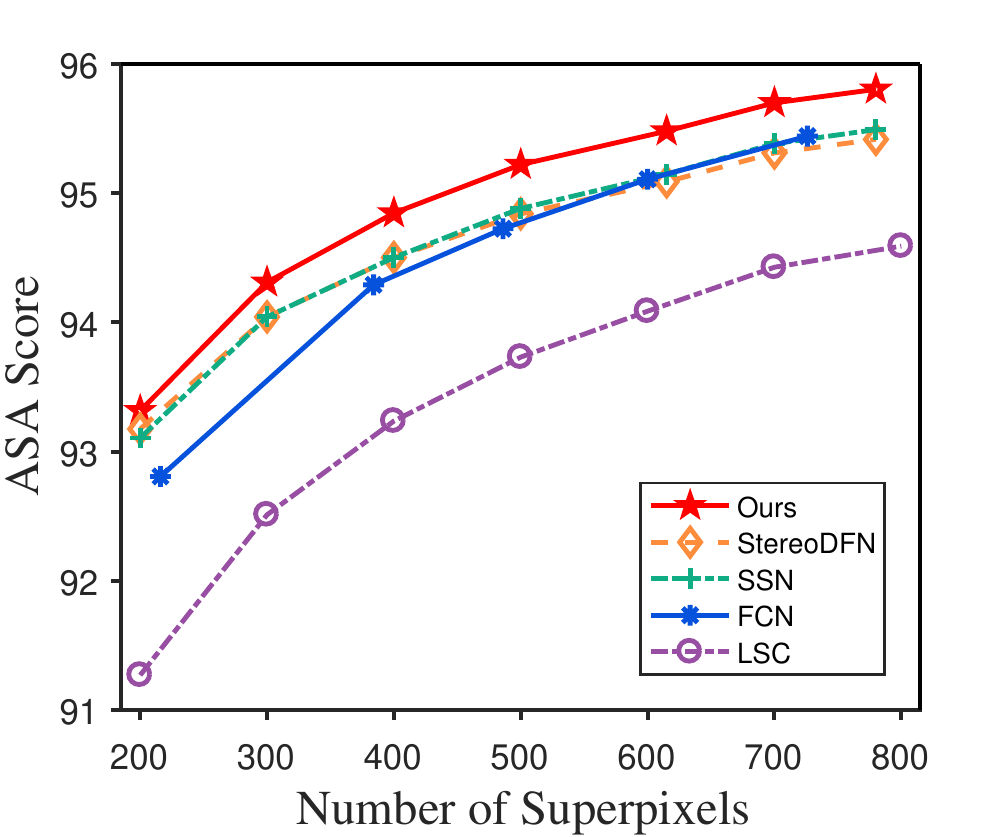}
        \includegraphics[width=.32\textwidth]{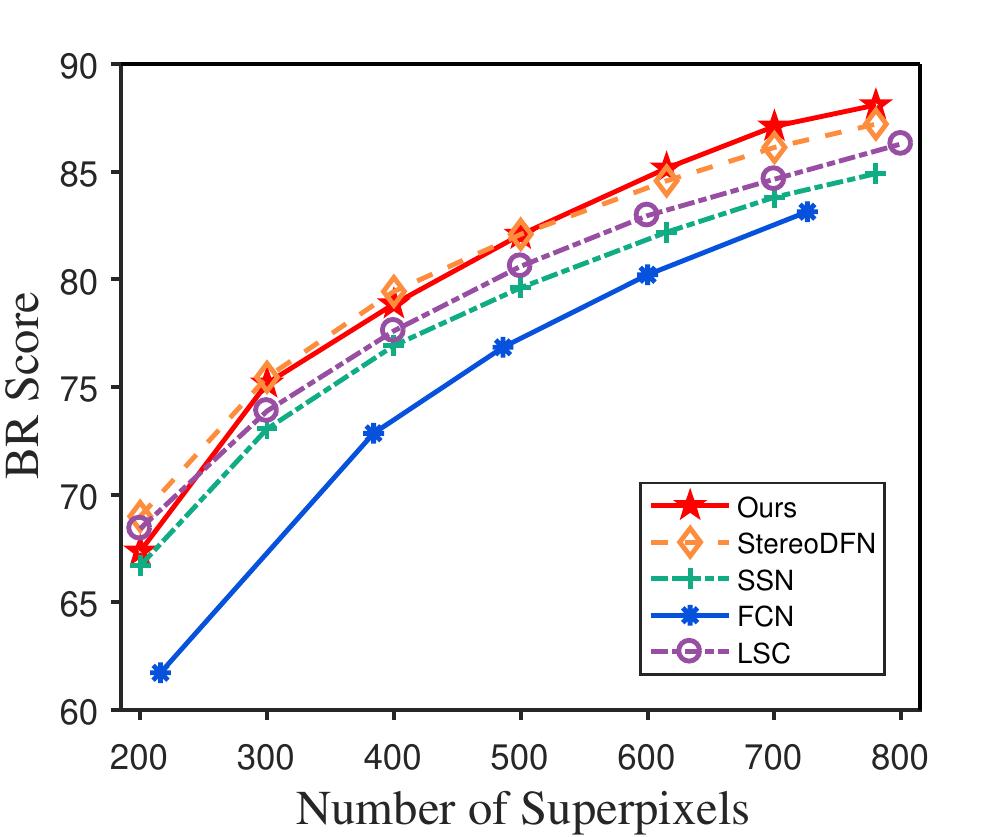}
        \includegraphics[width=.32\textwidth]{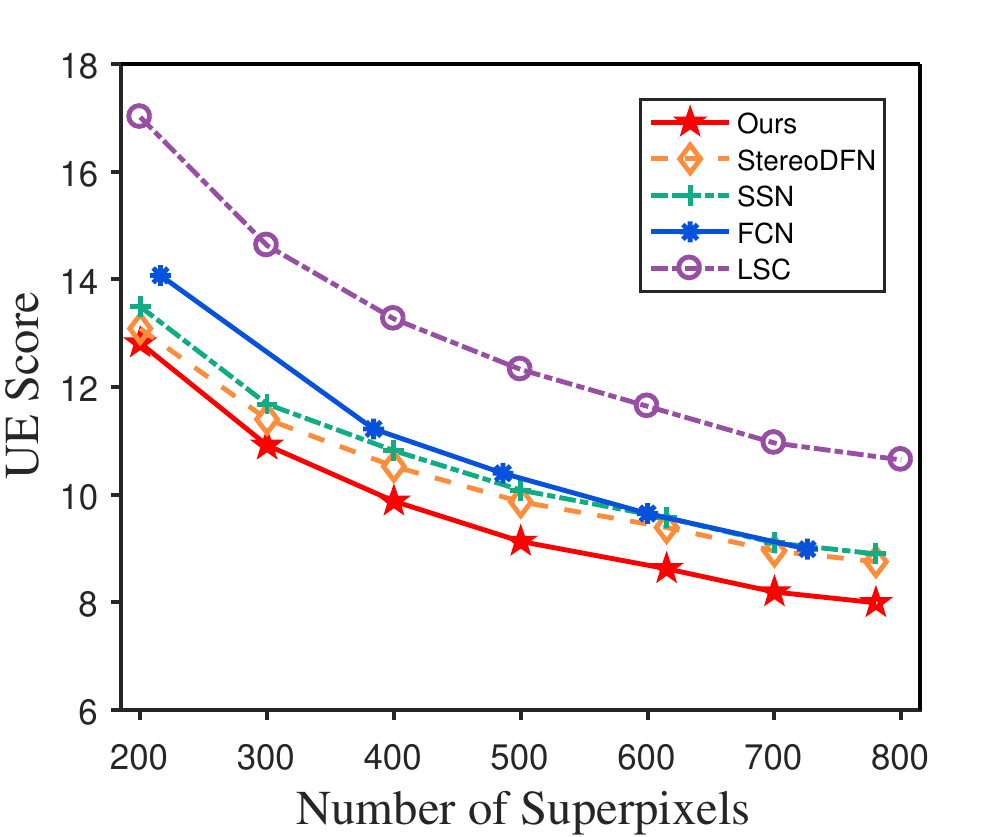}
        \label{fig:performance_a}
        }
    \subfigure[Performance on the validation set of Cityscapes dataset.]{
        \includegraphics[width=.32\textwidth]{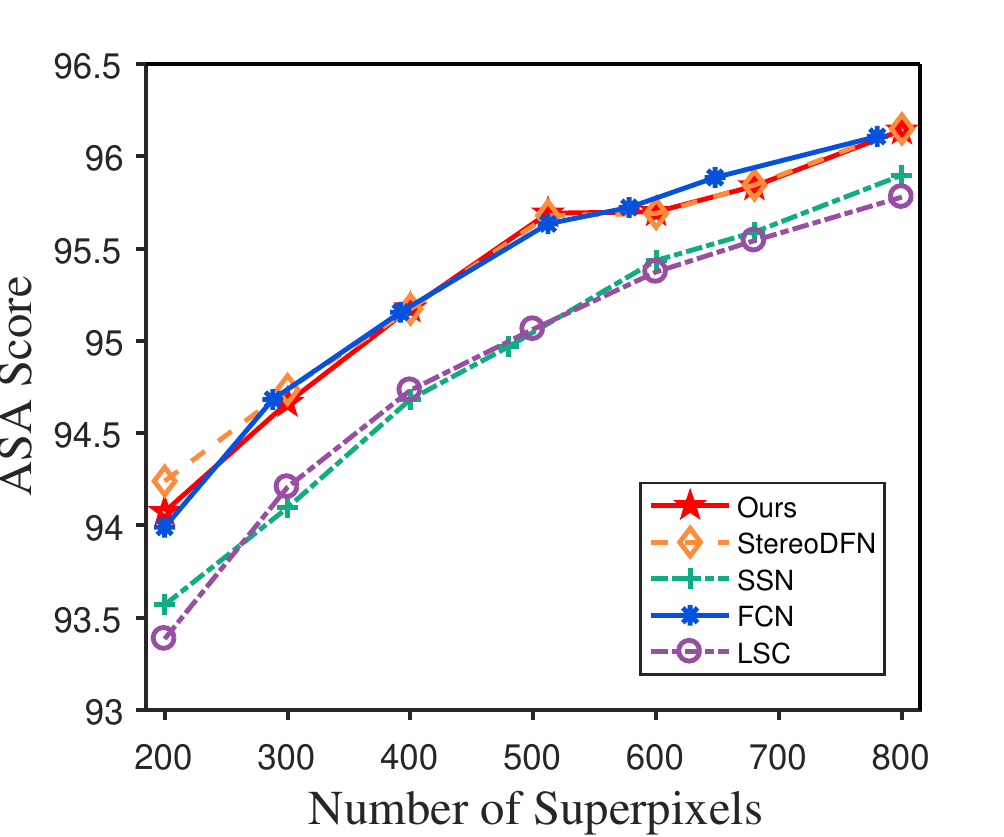}
        \includegraphics[width=.32\textwidth]{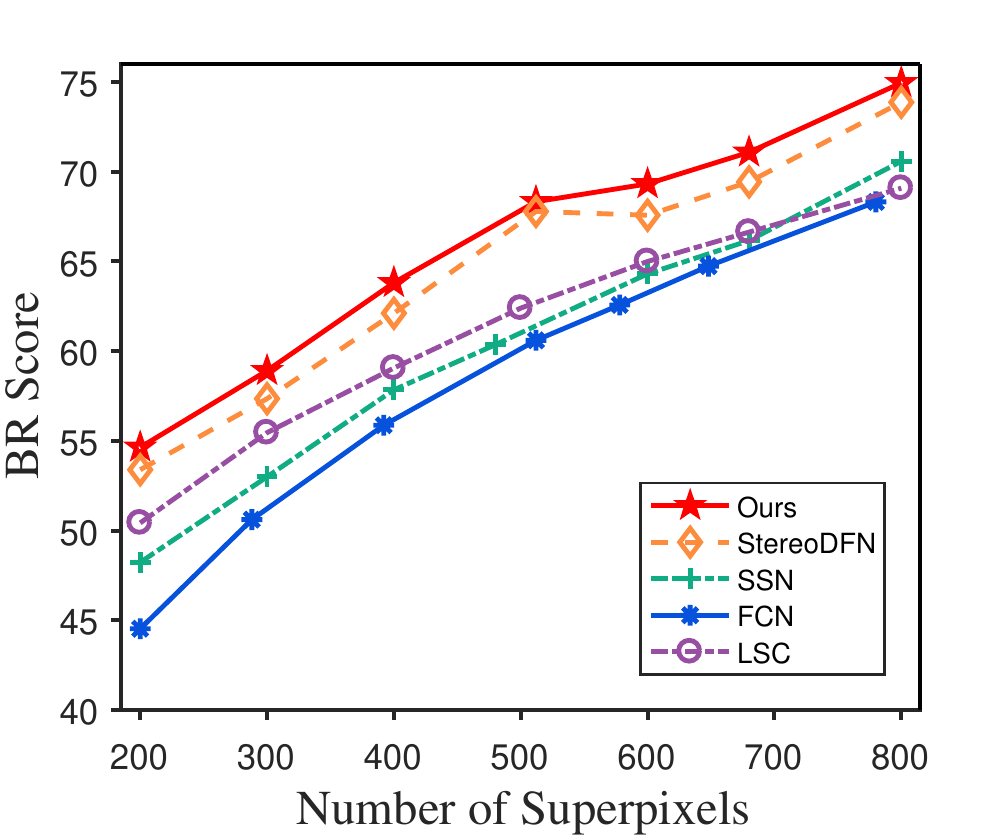}
        \includegraphics[width=.32\textwidth]{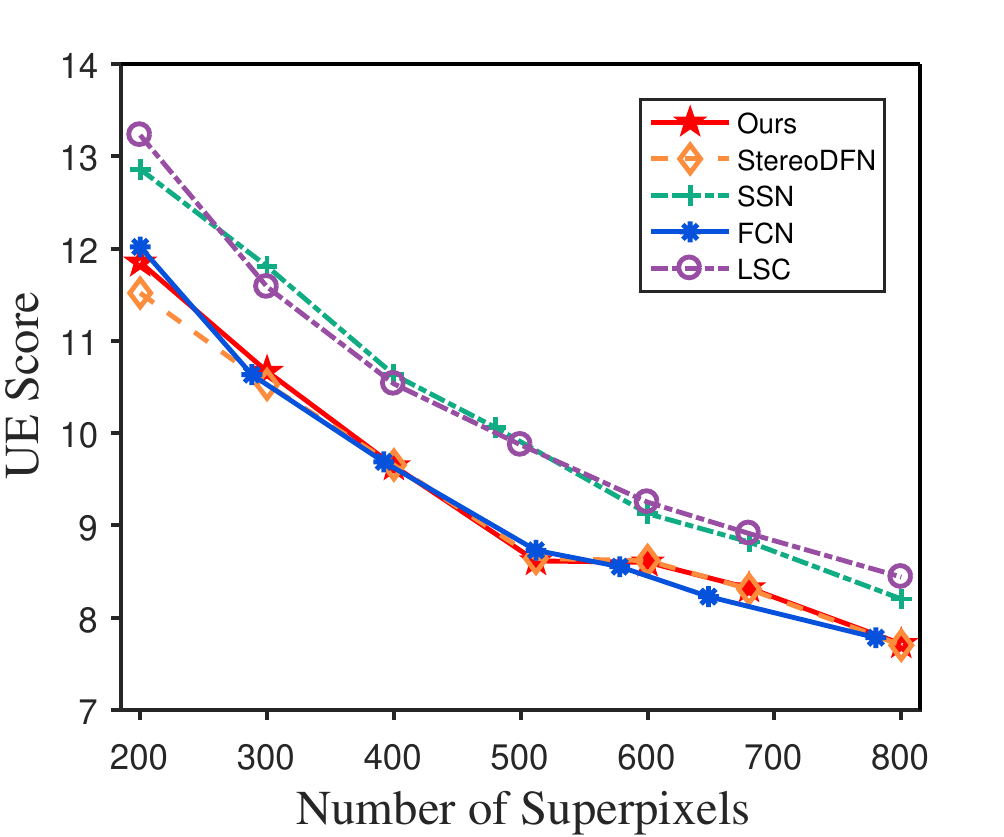}
        \label{fig:performance_b}
    }
    \caption{Quantitative comparison of the proposed method and other state-of-the-art methods.}
    \label{fig:performance}
\end{figure*}
%%%%%%%%%%%%%%%%%%%%%%%%%%%%%%%%%%%%%%%%%%%%%%%%%%%%%%%%%%%%%%%%%%%%%%%%%%
\textbf{Dynamic Fusion.}
Although spatial information is indispensable in superpixel segmentation task,
it does not always play the same important role of different regions in one image.
For example, for regions with sparse textures,
spatial information should be considered more to generate regular and compact superpixels.
On the other hand, for regions with dense edges and complex contents,
spatial information is relatively less important.
Therefore, to achieve consistently excellent performance in different conditions,
a dynamic fusion mechanism is designed to adaptively adjust the weighting of spatial information during fusion phase.

The dynamic fusion mechanism employs a channel-attention~\cite{senet} way
to adaptively aggregate and refine features.
More specifically, we first use a `Conv-BN-ReLU' block to fuse the features coarsely.
Then, a global average pooling layer with another `Conv-BN-ReLU' is followed to generate the global feature map.
Finally, a series operations are utilized to produce the weighting map,
which can be formulated as follows:
\begin{equation}
    \mathcal{W} = g \cdot \sigma(C(ReLU(LN(C(g))))),
\end{equation}
where $\mathcal{W}$ is the weighting map and $g$ is the global feature map.
$\sigma, C, ReLU$ and $LN$ represents sigmoid function, a convolution layer with $1 \times 1$ kernel size, ReLU activation function and layer-normalization, respectively.
%It amounts to feature selection and combination.
In this way, a more effective representation of spatial information with the guidance of weighting map can be obtained.
Finally, the weighting map is added to the input features and fed to the third `Conv-BN-ReLU' block to generate the adjusted features.
Fig.~\ref{fig:DF} presents the details of the Dynamic Fusion (DF) mechanism.

\subsection{Loss Functions}
We design two loss functions for optimizing our model.

\textbf{Semantic Loss.}
This function facilitates the superpixel adhere to semantic boundaries,
which utilize the cross-entropy loss function $SE$ to measure the loss:
\begin{equation}
    \mathcal{L}_{sem} = CE(S, S^*),
\end{equation}
where $S$ denotes the one-hot semantic label of ground truth and $S^*$ is the reconstructed semantic label.

\textbf{Stereo Loss.}
This loss function is designed to constrain the model to correctly estimate stereo correspondence.
We also add valid mask to eliminate the problems caused by occlusion.
Stereo loss is defined as:
\begin{equation}
\begin{aligned}
    \mathcal{L}_{stereo} =& \left\|O_{L \to R} \circ (I_{L} - \mathcal{M}_{R \to L} \otimes I_{R})\right\|_1+
	\\ &\left\|O_{R \to L} \circ (I_{R} - \mathcal{M}_{L \to R} \otimes I_{L})\right\|_1,
\end{aligned}
\end{equation}
where $I_L,~I_R$ denotes the left and right image, respectively. $\circ$ denotes Hadamard product.

The total loss is the sum of these two functions:
\begin{equation}
    \mathcal{L}_{total} = \mathcal{L}_{sem} + \lambda \mathcal{L}_{stereo},
\end{equation}
where $\lambda$ is empirically set to $1.0$ for balancing the scales of different losses.

All above are the basic definitions of the metrics we used for evaluating,
more details can be seen in \cite{benchmark}.

\section{Experiments and results}

\subsection{Experimental Setup}
\textbf{Implementation Details.} 
We apply a batch-mode learning method with a batch size of $8$ 
to train our model for 20K iterations.
The Adam with default parameters ($\beta_1 = 0.9,~\beta_2=0.999$) is utilized to optimize the network.
In addition, the initial learning rate is $2\times10^{-4}$ and decreases by half every 2k iterations.
After 8k iterations, the learning rate is fixed to $2\times10^{-5}$.
During training phase, we randomly crop the images into size $200 \times 200$ for augmenting the training data.
Following~\cite{SLIC,SSN}, 
stereo image pairs with Lab color space is used as input,
and the Lab color space information is zoomed by multiplying a coefficient $\beta = \eta \max(m_w/n_w,m_h/n_h)$, 
where $m$ and $n$ represent the number of superpixels and pixels, $\eta$ is equal to $2.5$.
All experiments are implemented by PyTorch framework on a PC with NVIDIA RTX A4000 GPU.

%%%%%%%%%%%%%%%%%%%%%%%%%% Figure: visual_img_compare KITTI%%%%%%%%%%%%%%%%%%%%%%%%%%
\begin{figure*}[t!]
\centering
    \includegraphics[width=\linewidth]{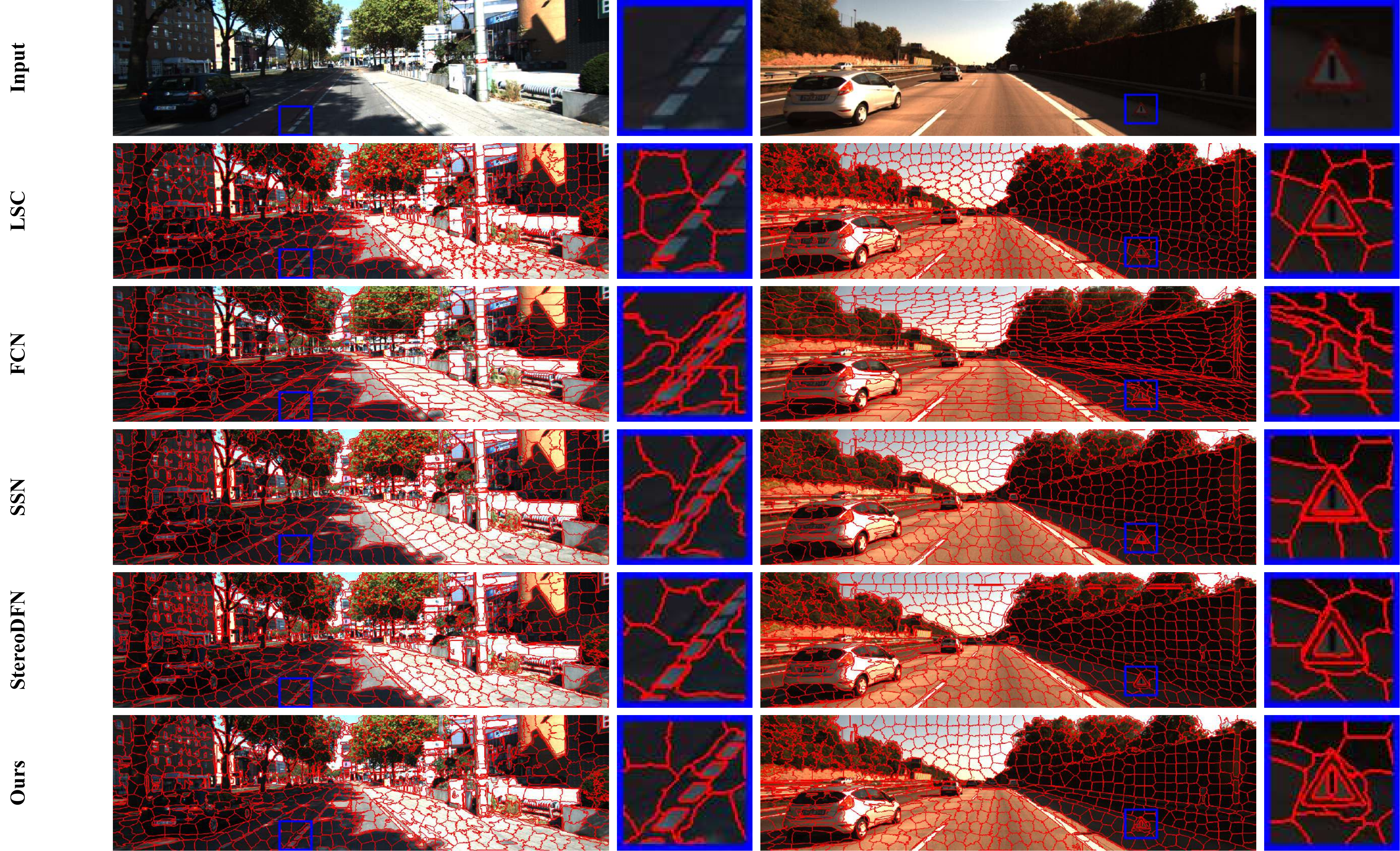}
    \caption{Qualitative comparison of the proposed method and other state-of-the-art methods in KITTI2015.}
    \label{fig:KITTI_Compare}
\end{figure*}
\textbf{Datasets.}
Following the experiment settings in \cite{2021Stereo}, we use KITTI2015~\cite{KITTI} and Cityscapes\cite{Cityscapes} datasets to train and test our model.
KITTI2015 contains 200 stereo image pairs with semantic annotations of left images,
we select 150 for training and 50 for testing.
Moreover, to further indicate the superiority of the proposed method,
we also use the Cityscapes dataset for evaluation.
Cityscapes is a larger and more challenging dataset,
which contains extensive stereo image pairs captured with diverse scenes, weathers and illumination conditions.
% The format of semantic annotations is same as KITTI2015.
Since the test set of Cityscapes is not public available,
we use the validation set for comparing,
which is consist of 500 stereo images.
Furthermore, the image of Cityscapes has been scaled to quarter-resolution for convenience.

\textbf{Evaluation Metrics.}
In our experiments, we use three widely used metrics to evaluate the performance of our model,
which are achievable segmentation accuracy (ASA), 
undersegmentation error (UE),
and boundary recall (BR).
For superpixel map $S=\{S_i\}$ and ground truth of semantic label $G=\{G_j\}$, 
The detailed definitions of these metrics are as follows:

\textit{Achievable segmentation accuracy~(ASA):} 
ASA is a metric for evaluating the upper bound on the achievable segmentation accuracy,
which can be formulated as:
\begin{equation}
    ASA(S, G) = \frac{\sum_i max_j|S_i \cap G_j|}{\sum_j |G_j|}.
\end{equation}

\textit{Undersegmentaion error~(UE):}
UE essentially measures the error of superpixel segmentation with respect to the ground truth.
The UE is defined as:
\begin{equation}
    UE(S, G) = \frac{1}{|G|}\sum_{G_j}\frac{(\sum_{S_i \cap G_j}|S_i|) - |G_j|}{|G_j|}.
\end{equation}

\textit{Boundary recall~(BR):}
This is a metric of how well the superpixel adhere to image boundaries.
We use a coefficient $r$ to divide all pixels into two categories.
$TP(S,G)$ is the boundary pixels in $G$ for which there is a boundary pixel in $S$ in range $r$,
and $FN(S,G)$ is the opposite of it.
Then BR can be formulated as:
\begin{equation}
    BR(S, G) = \frac{TP(S, G)}{TP(S, G) + FN(S, G)}.
\end{equation}
\begin{figure*}[t]
    \centering
    \includegraphics[width=\linewidth]{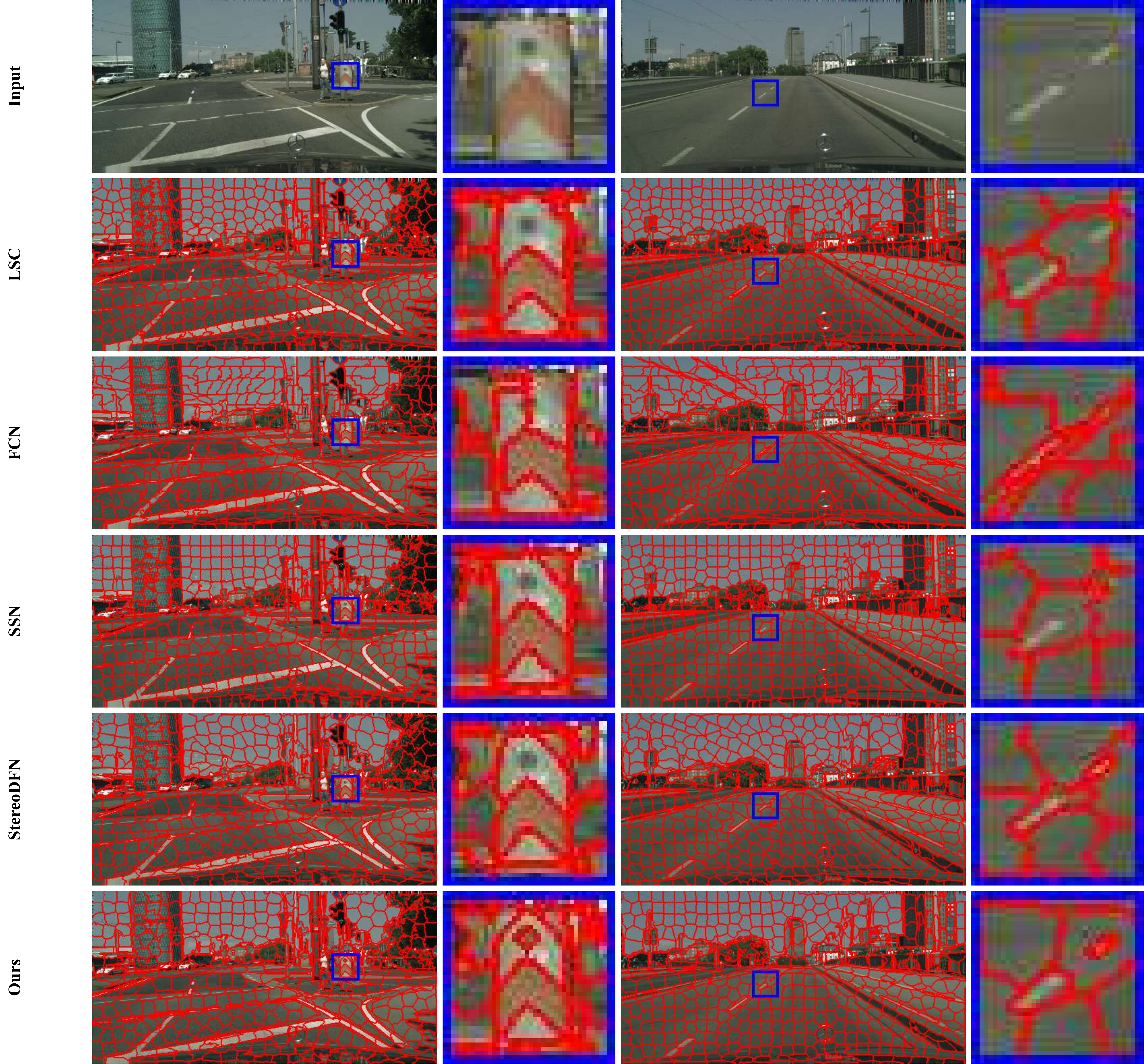}
    \caption{Qualitative comparison of the proposed method and other state-of-the-art methods in Cityscapes.}
    \label{fig:Cityscapes_Compare}
\end{figure*}
\subsection{Comparison with State-of-the-Arts}
In this part, we compare our method with some state-of-the-art methods, 
including SSN\cite{SSN}, FCN\cite{FCN}, LSC\cite{LSC}, StereoDFN\cite{2021Stereo}.
All of the compared methods are adopted the parameters setting of original works and implemented by the code released by~\cite{benchmark} or the original authors.

\textbf{Quantitative Comparison.}
Fig.~\ref{fig:performance_a} and Fig.~\ref{fig:performance_b} shows the quantitative comparison results of our proposed method and other state-of-the-art methods on KITTI2015 and Cityscapes, respectively.
We can see that our method achieves the top score on KITTI2015 and comparaZble performance to FCN and StereoDFN on Cityscapes.
Taking 700 superpixels for example, in terms of ASA, UE and BR, our method achieves the minimum percentage gain (computed with the highest score of other methods) on KITTI2015 is 0.4$\%$, 9.3$\%$, 1.1$\%$, while the maximum percentage gain is 1.4$\%$, 33.8$\%$, 4.8$\%$, respectively.
On Cityscapes, our method achieves minimum and maximum percentage gain of BR is 2.4$\%$ and 9.8$\%$, respectively, and also achieves a comparable performance to FCN and StereoDFN in terms of ASA and UE.

\textbf{Qualitative Comparison.}
As the qualitative comparison results shown in Fig.~\ref{fig:KITTI_Compare} and Fig.~\ref{fig:Cityscapes_Compare},
it is clear that our method achieves the best visual performance since it can adhere to object boundaries and preserve texture better.
More specifically, on KITTI2015, we can see that our method can adhere to the boundary of various lane lines more accurately and capture the detail of the warning sign in low light while other methods cannot.
For Cityscapes, only our method can capture the details on the warning sign and green light on the traffic light while adhering the image boundaries well.

In conclusion, through the quantitative comparison results based on standard evaluation criteria, we can see that our method outperforms other methods in most cases, and achieves the best visual performance in qualitative comparison.
The impressive performance of our method also verifies the superiority of the proposed spatial decoupling mechanism.

\subsection{Ablation Experiments}
In order to validate the effectiveness of each component in our proposed network,
we perform extensive ablation experiments on KITTI2015.
There are three types of ablation models including T0, T1 and T2,
T0 denotes the full model.
For baselines in T1,
they can indicate our DSEM can combine the spatial information better,
while baselines in T2 show that our method makes a good use of information from another viewpoint and improve the performance of superpixel segmentation.
In addition, T1 and T2 contain 3 ablation models respectively,
More specifically,
\begin{itemize}
    \item B1 denotes the ablation model without SE and DF modules.
    \item B2 stands for the ablation model with spatial information (XY) without SE and DF modules.
    \item B3 represents the ablation model without DF module.
    \item B4 means that the ablation model without stereo loss, and does not consider stereo features alignment and occlusion problem.
    \item B5 refers to the ablation model without considering occlusion problem.
    \item B6 is the ablation model without stereo loss
\end{itemize}

All of the ablation experiments are trained for 20K iterations.
The specific structure of each ablation model also has been shown in TABLE~\ref{table:ablation experiments}.
%%%%%%%%%%%%%%%%%%%%%%%%%%%%%%%%%%%%%%%%%%%%%%%%%%%%%%%%%%%%%%%%%%%%%%%%%%
%%%%%%%%%%%%%%%%%%%%%%%%%%%%%%%%%%%%%%%%%%%%%%%%%%%%%%%%%%%%%%%%%

\begin{figure*}[h!]
\centering
    \includegraphics[width=\linewidth]{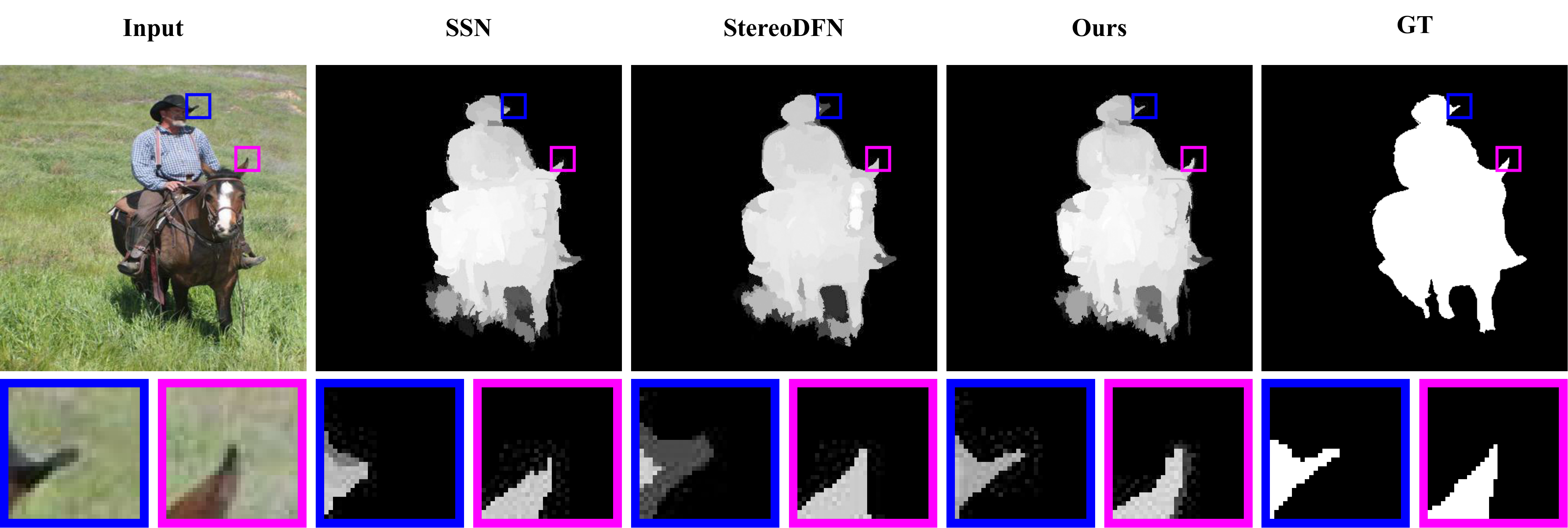}
    \caption{Visual comparison of SOD results with different superpixel segmentation methods. Note that our method can preserve more details than others.}
    \label{fig:SOD_Compare}
\end{figure*}
%%%%%%%%%%%%%%%%%%%%%%%%%%%%%%%%%%%%%%%%%%%%%%%%%%%%%%%%%%%%%%%%%%%%%%%%%%
%%%%%%%%%%%%%%%%%%%%%%%%% Ablation Table %%%%%%%%%%%%%%%%%%%%%%%%%
\begin{table}[h]
\centering
\begin{tabular}{c|c|c|cccc|c}
\toprule
{\multirow{2}{*}{Type}} & {\multirow{2}{*}{ID}} & {\multirow{2}{*}{Input}} & \multicolumn{4}{c|}{Component} & \multirow{2}{*}{Stereo Loss} \\ \cline{4-7}
  &               &           & \multicolumn{1}{c|}{SFA} & \multicolumn{1}{c|}{OH} & \multicolumn{1}{c|}{SE} & \multicolumn{1}{c|}{DF} & \\ \hline
T0 & B0 & Stereo    & \checkmark & \checkmark & \checkmark & \checkmark & \checkmark \\ \hline
  & B1 & Stereo    & \checkmark & \checkmark &            &            & \checkmark \\ 
T1 & B2 & Stereo+XY & \checkmark & \checkmark &            &            & \checkmark \\ 
  & B3 & Stereo    & \checkmark & \checkmark & \checkmark &            & \checkmark \\ \hline
  & B4 & Single    &            &            & \checkmark & \checkmark &            \\
T2 & B5 & Stereo    & \checkmark &            & \checkmark & \checkmark & \checkmark \\ 
  & B6 & Stereo    & \checkmark & \checkmark & \checkmark & \checkmark &            \\ \bottomrule
\end{tabular}

\caption{Detailed setup for ablation experiments.
         B$n$ and T$n$ denote 
         the $n$th baseline and the $n$th type respectively.}
\label{table:ablation experiments}
\end{table}
\begin{figure}[h]
    \centering
    \subfigure{
        \includegraphics[width=.935\linewidth]{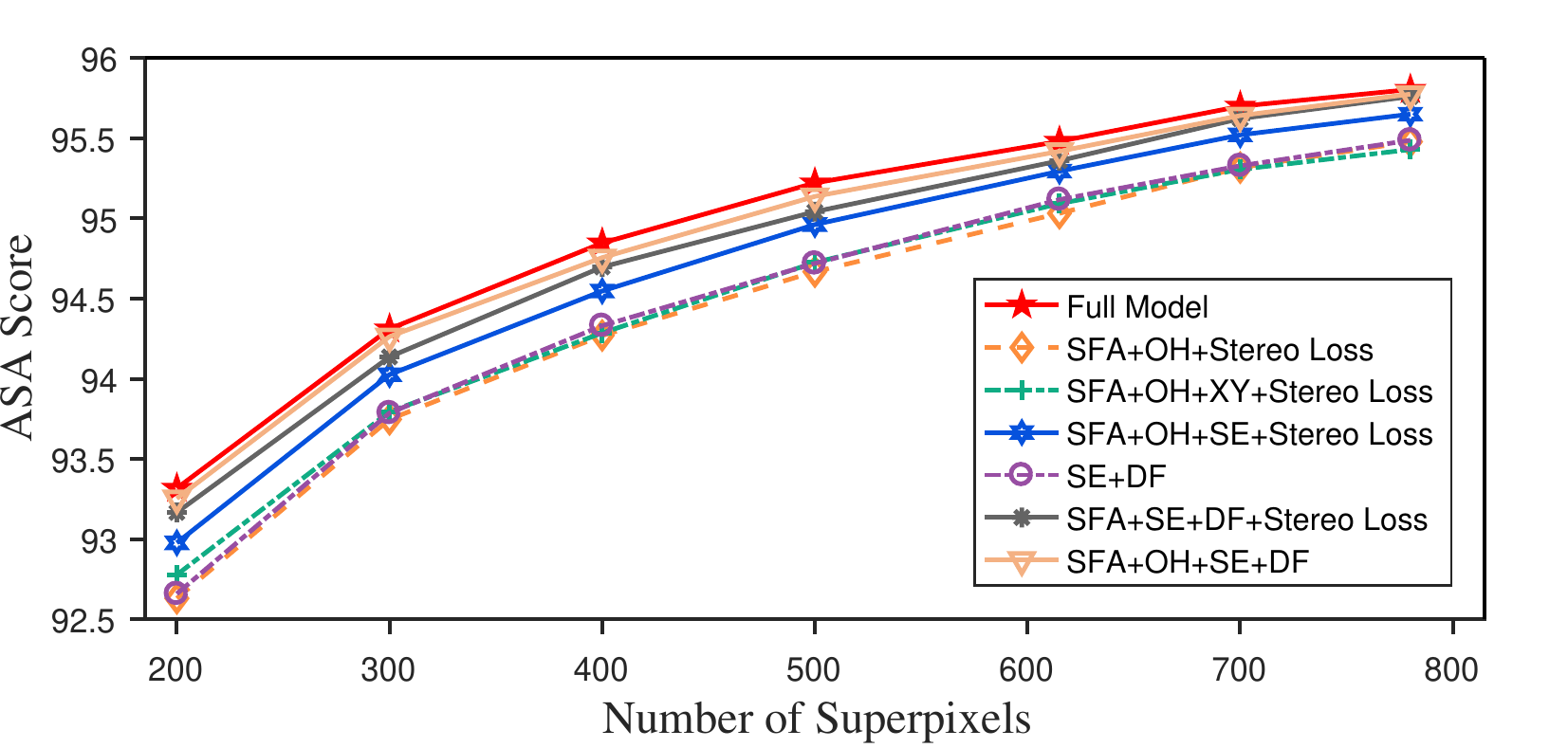}
	}
	\subfigure{
		\includegraphics[width=.935\linewidth]{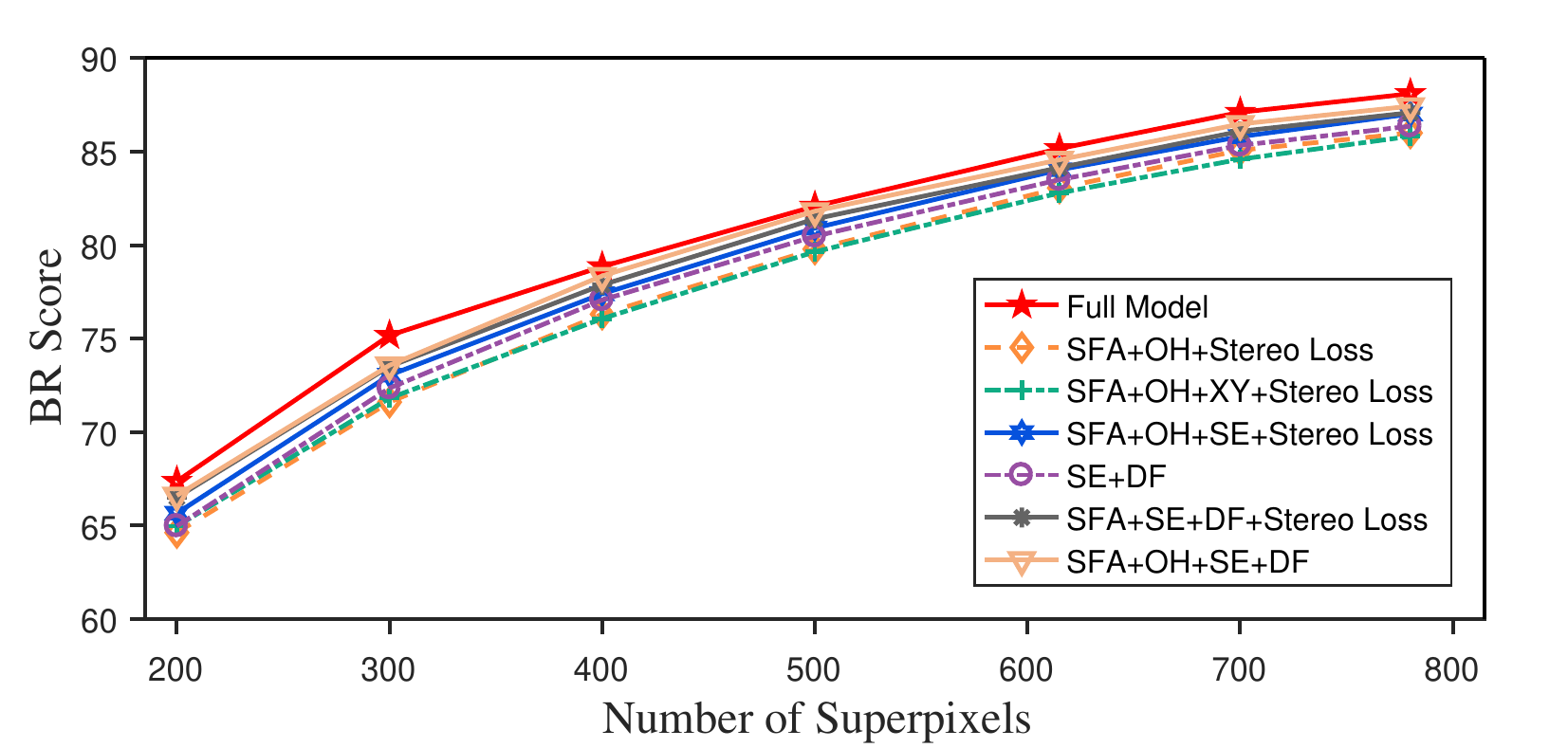}
	}
    \caption{\textbf{Ablation studies on KITTI2015.} The top figure shows the contributions of each component through ASA score, while the bottom figure through BR score.}
    \label{fig:ablation}
\end{figure}
\textbf{Effectiveness of Each Component.}
Fig.~\ref{fig:ablation} reports the quantitative comparison results of ablation models on KITTI2015.
We can see that adding spatial information directly can not make full use of it.
However,
the spatial information can be embedded into the network better through our SE and DF modules, resulting in higher performance gains.
In addition, 
the DSFM module and Stereo Loss also play an important role, which can solve the stereo features alignment and occlusion problem and constrain the model to correctly model stereo correspondence, respectively.

\textbf{Influence of Spatial Information.}
From Fig.~\ref{fig:ablation},
we can observe that model with SE module tends to have a larger performance improvement than the model without SE module,
which proves that the spatial information is helpful to generate regular and compact superpixels.
Furthermore, adjusting the weighting of the spatial information adaptively through the DF module can make better use of it to further improve the performance.

\section{Application on Salient Object Detection}
Salient object detection (SOD) has attracted increasing interest in recent years,
since it plays a significant role in many popular computer vision tasks,
including object recognition and detection
% \cite{ren2013region,zhang2017bridging},
\cite{ren2013region,zhang2017bridging},
% image retargeting \cite{sun2011scale,ding2011importance},
image retargeting \cite{ding2011importance,sun2011scale},
semantic segmentation \cite{wei2016stc,wang2018weakly}, etc.
To improve the performance of salient object detection,
Zhu \textit{et al.} \cite{zhu2014saliency} propose an superpixel-based salient object detection method,
they treat the saliency object detection problem as a saliency value optimization problem for all superpixels in an image.
Moreover, they observe that background regions are more connected to image boundaries than salient object regions.
Therefore, they propose a measure called boundary connectivity,
which is utilized to perform salient object detection in their proposed method.
The boundary connectivity is defined as follows:
\begin{equation}
    \label{eq:bndcon}
    BndCon(R)=\frac{|\{p \mid p \in R, p \in B n d\}|}{\sqrt{|\{p \mid p \in R\}|}}, 
\end{equation}
where $p$ is a patch of an image and $Bnd$ is the set of all image boundary patches.
% For the reason that Eq.~(\ref{eq:bndcon}) is difficult to compute directly,
% since hard segmentation will introduce discontinuous artifacts,
% Zhu \textit{et al.} \cite{zhu2014saliency} propose a “soft” approach based on superpixel segmentation instead of hard segmentation,
% which means that the definition of boundary connectivity similar to Eq.~(\ref{eq:bndcon}) can be formulated as Eq.~(\ref{eq:bndcon_pro}):
% \begin{equation}
%     \label{eq:bndcon_pro}
%     BndCon(p)=\frac{Len_{b n d}(p)}{\sqrt{Area(p)}}, 
% \end{equation}
% where $Area$ is the “spanning area” of each superpixel $p$ and $Len_{b n d}$ is the length along the boundary.

\begin{table}[t]
\normalsize
    \centering
    \begin{tabular}{p{2cm}p{1.2cm}<{\centering}p{2.2cm}<{\centering}p{1.2cm}<{\centering}}
     \toprule
     Method & SSN & StereoDFN & Ours\\
     \midrule
     MAE $\downarrow$ & 0.1783& 0.1794 & \textbf{0.1777}\\
     
     E-measure $\uparrow$ & 0.6489& 0.6427 & \textbf{0.6498}\\
     \bottomrule
    \end{tabular}
    \caption{Results on NJU2K benchmark. $\uparrow$ denotes the higher is the better, and $\downarrow$ is contrary.}
\label{table:application}
\end{table}
To indicate our method can perform better in downstream task,
we use three state-of-the-art methods, which are our proposed method, StereoDFN\cite{2021Stereo} and SSN\cite{SSN}
to replace the default SLIC\cite{SLIC} as the superpixel segmentation method of \cite{zhu2014saliency}.
In our experiments, we use NJU2K \cite{ju2015depth} for evaluation.
NJU2K is a large dataset widely used for salient object detection of stereo images, which contains 2000 stereo image pairs, involving various objects and scenarios of different difficulty levels.
Moreover, we select the first 400 images of NJU2K and resize all of them to size $400 \times 400$ for the ease of experimentation.

Following \cite{zhu2014saliency}, we choose the mean absolute error (MAE) \cite{perazzi2012saliency} to evaluate each method quantitatively, which is a metric to measure the average difference between the binary ground truth and saliency prediction map.
However, MAE only focus on pixel-wise error.
To consider structure cues, we also introduce Enhanced-alignment measure (E-measure) \cite{fan2018enhanced} as our evaluation metric.
    
The results of the quantitative evaluation are shown in Table \ref{table:application},
we can see that our method achieves the best performance in terms of MAE and E-measure.
In addition, the visual comparison results in Fig.~\ref{fig:SOD_Compare} also illustrate the saliency map generated based on our method can focus on more details than other state-of-the-art methods, which validates that our method can perform well in downstream task both qualitatively and quantitatively.

\section{Conclusion}
Previously, stereo superpixel segmentation methods neglect the coupling of stereo features and spatial features, which may impose strong constraints on spatial information while modeling the correspondence between stereo image pairs.
To solve this problem, we have presented an end-to-end stereo superpixel segmentation network with a decoupling mechanism of spatial information to eliminate such negative influence.
In addition, spatial information is adjusted adaptively through our dynamic fusion mechanism in dynamic spatiality embedding module to generate regular and compact superpixels.
Extensive experiments on several popular datasets have shown that our proposed method achieves the state-of-the-art performance and performs well in downstream task.
The effectiveness of the components handling the spatial information and stereo features have also been verified in our ablation studies.

%\section*{Acknowledgment}


\ifCLASSOPTIONcaptionsoff
  \newpage
\fi



\bibliographystyle{IEEEtran}
\bibliography{stereo_superpixel_TMM}
\end{document}